\documentclass[sigconf]{acmart}

\usepackage[caption=false]{subfig}
\usepackage[export]{adjustbox}
\usepackage{makecell}
\usepackage{bm}
\usepackage{titlesec}
\usepackage{amsmath,amsfonts}
\usepackage{algorithmic}
\usepackage{graphicx}
\usepackage{textcomp}
\usepackage{xcolor}
\usepackage{multirow}
\usepackage{hyperref}
\usepackage{booktabs}
\usepackage{hhline}
\usepackage[shortlabels]{enumitem}
\usepackage{soul}
\usepackage{fancyhdr}   % custom header/footer
\usepackage{calc}
\usepackage{atbegshi}
\usepackage{tikz}

\usepackage{geometry}
\usepackage[utf8]{inputenc}
\usepackage{color, colortbl}  % For highlighting

\titlespacing*{\section}{0pt}{4pt}{1pt}
\titlespacing*{\subsection}{0pt}{3pt}{-1pt}
\titlespacing*{\subsubsection}{0pt}{1pt}{4pt}

\AtBeginDocument{%
  }

\setcopyright{acmlicensed}
\copyrightyear{2025}
\acmYear{2025}
\definecolor{rubinered}{HTML}{CE0058}

\definecolor{green}{rgb}{0.0, 0.65, 0.31}

\definecolor{bleudefrance}{rgb}{0.19, 0.55, 0.91}

\definecolor{ao(english)}{rgb}{0.0, 0.5, 0.0}

\definecolor{violet}{HTML}{6a51a3}

\begin{document}

% \title{Wheelchair Alternate Data}
\title{Scaling Human Activity Recognition: A Comparative Evaluation of Synthetic Data Generation and Augmentation Techniques}

\author{Zikang Leng}
\email{zleng7@gatech.edu}
\orcid{0000-0001-6789-4780}
\affiliation{%
  \institution{Georgia Institute of Technology}
  \city{Atlanta, Georgia}
  \country{USA}}

\author{Archith Iyer}
\email{aiyer351@gatech.edu}
\orcid{0009-0002-3559-6038}
\affiliation{%
  \institution{Georgia Institute of Technology}
  \city{Atlanta, Georgia}
  \country{USA}}

\author{Thomas Plötz}
\email{thomas.ploetz@gatech.edu}
\orcid{0000-0002-1243-7563}
\affiliation{%
  \institution{Georgia Institute of Technology}
  \city{Atlanta, Georgia}
  \country{USA}}

\renewcommand{\shortauthors}{ Zikang Leng, Archith Iyer, \& Thomas Plötz}

\begin{abstract}

Human activity recognition (HAR) is often limited by the scarcity of labeled datasets due to the high cost and complexity of real-world data collection. To mitigate this, recent work has explored generating virtual inertial measurement unit (IMU) data via cross-modality transfer. While video-based and language-based pipelines have each shown promise, they differ in assumptions and computational cost. Moreover, their effectiveness relative to traditional sensor-level data augmentation remains unclear. In this paper, we present a direct comparison between these two virtual IMU generation approaches against classical data augmentation techniques.
We construct a large-scale virtual IMU dataset spanning 100 diverse activities from Kinetics-400 and simulate sensor signals at 22 body locations. The three data generation strategies are evaluated on benchmark HAR datasets (UTD-MHAD, PAMAP2, HAD-AW) using four popular models. Results show that virtual IMU data significantly improves performance over real or augmented data alone, particularly under limited-data conditions. We offer practical guidance on choosing  data generation strategies and highlight the distinct advantages and disadvantages of each approach.

\end{abstract}

% \iffalse
\begin{CCSXML}
<ccs2012>
<concept>
<concept_id>10003120.10003138</concept_id>
<concept_desc>Human-centered computing~Ubiquitous and mobile computing</concept_desc>
<concept_significance>500</concept_significance>
</concept>
<concept>
<concept_id>10010147.10010178</concept_id>
<concept_desc>Computing methodologies~Artificial intelligence</concept_desc>
<concept_significance>500</concept_significance>
</concept>
</ccs2012>
\end{CCSXML}

\ccsdesc[500]{Human-centered computing~Ubiquitous and mobile computing}
\ccsdesc[500]{Computing methodologies~Artificial intelligence}
% \fi 

\keywords{Virtual IMU Data; Activity Recognition; Data Augmentation
}

\maketitle

\iffalse
%%% custom footer, remove for submission ----------
\pagestyle{fancy}
\fancyhf{}
\renewcommand{\headrulewidth}{0pt}
\AtBeginShipout{\AtBeginShipoutAddToBox{%
  \begin{tikzpicture}[remember picture, overlay, red]
    \node[anchor=south, font=\LARGE] at ([yshift=15mm]current page.south) {This manuscript is under review. Please write to zleng7@gatech.edu for up-to-date information};
  \end{tikzpicture}%
}}
%---------------------------------------------------
%---------------------------------------------------
\fi

\section{Introduction}

A persistent challenge in Human activity recognition (HAR) is the lack of large-scale, richly annotated IMU datasets. Collecting such data is costly and labor-intensive, requiring precise sensor placement, diverse participant recruitment, and manual labeling—often resulting in datasets with limited activity coverage and constrained sensor setups \cite{chen2021sensecollect}. Consequently, models trained on these datasets tend to generalize poorly across users and environments, hindering real-world deployment \cite{Plotz2023IfOnlyWe}.

To address the scarcity of labeled IMU datasets, recent work has explored cross-modality transfer techniques that generate virtual IMU data from more abundant sources such as video and text. Video-based methods like IMUTube \cite{kwon2020imutube, kwon2021approaching} transform 2D RGB activity videos—sourced from platforms like YouTube or datasets such as Kinetics-400 \cite{kay2017kineticshumanactionvideo}—into virtual IMU signals through pose estimation and biomechanical signal modeling. In contrast, text-based pipelines such as IMUGPT \cite{leng2023generating, leng2024imugpt} leverage large language models (LLMs) and text-to-motion models like T2M-GPT \cite{zhang2023generating} to synthesize human motion from natural language descriptions, producing semantically diverse virtual IMU data. While both approaches offer scalability far beyond conventional IMU datasets, they differ in their underlying assumptions and computational requirements.

Cross-modality transfer approaches can be understood as an alternative form of data augmentation that synthesizes new training examples from external modalities—such as text or video—rather than modifying existing IMU signals. By generating virtual IMU data that reflects diverse activity variations, these methods enrich the training data and can boost downstream HAR performance when used alongside real-world data. While both text-based and video-based pipelines have individually demonstrated benefits, they have only been evaluated in isolation. Currently, no study has directly compared these two approaches. Moreover, despite their shared objective with traditional sensor-level data augmentation, cross-modality methods have not been systematically benchmarked against them. This lack of comparative analysis leaves practitioners with limited guidance on how to choose between these strategies for improving HAR models.

In this paper, we compare these three methods for large-scale IMU data generation. Our results show that incorporating virtual IMU data consistently enhances model performance compared to training on real data alone or with traditional data augmentation—particularly when real data is scarce. Video-generated virtual IMU data generally provide more accurate motion information, while text-generated virtual IMU data contribute semantic and contextual diversity; combining both leads to the model with the best performance. Additionally, we find that virtual IMU data greatly benefit underrepresented activity classes and can help mitigate class imbalance. Our findings highlight the practical advantages of cross-modality transfer approaches and offer insights into how their complementary strengths can be leveraged for scalable HAR.

Our key contributions are: 

\vspace*{-1em}
\begin{enumerate}[label=(\arabic*)]
    \item We generate 
    %\thomas{Develeop?} \zikang{now changed} 
    a large-scale virtual IMU dataset covering 100 diverse human activities from Kinetics-400, with simulated signals at \emph{22 distinct sensor locations}. This dataset will be made publicly available upon acceptance.
    
    \item We evaluate three data generation methods—classical sensor-level augmentation, video-based virtual IMU data generation using IMUTube, and text-based virtual IMU data generation using IMUGPT—on three HAR datasets: PAMAP2 \cite{Reisspamap}, HAD-AW \cite{Mohammed2018hadaw}, and UTD-MHAD \cite{chen2015mhad}.
    
    \item We provide practical guidelines for researchers and practitioners on selecting the most appropriate data generation technique based on resource constraints, desired activity complexity, and deployment scenarios.
\end{enumerate}

\section{Related Work}

\subsection{Human Activity Recognition}

HAR has traditionally relied on the Activity Recognition Chain (ARC), which includes recording, signal processing, segmentation, feature extraction, and classification. Early methods employed classical machine learning models trained on handcrafted features derived from time- and frequency-domain statistics.

With the rise of deep learning, end-to-end models operating directly on raw sensor signals have become prevalent. Convolutional neural networks (CNNs) \cite{munzner2017cnn}, long short-term memory networks (LSTMs) \cite{chen2016lstm}, and hybrid architectures such as DeepConvLSTM \cite{convlstm, singh2021DeepConvLSTM}, BiLSTM with attention \cite{zhang2023bilstm}, and transformer-based models \cite{Dirgová2022transform} achieve state-of-the-art results by capturing local temporal dynamics, long-range dependencies, and inter-channel relationships.

However, these supervised DL approaches require substantial labeled data to perform effectively \cite{Plotz2023IfOnlyWe}. To mitigate this dependency, alternative learning paradigms have emerged. Self-supervised learning (SSL) frameworks pre-train models using auxiliary objectives— such as contrastive prediction or temporal ordering—before fine-tuning on limited labeled data \cite{chen2024utilizing, chen2024enhancing, Haresamudram2022ssl, oord2018representation}. Transfer learning further extends model generalization by adapting representations learned in one domain or user group to new contexts \cite{an2023transfer, dhekane2024transfer, soleimani2021cross}.

Building upon these directions, our work investigates the effectiveness of cross-modality transfer techniques to generate virtual IMU data from alternative modalities. 

\subsection{Cross-Modality Transfer}

Cross-modality transfer methods aim to address the scarcity of labeled IMU datasets by leveraging existing annotated datasets from other sensing modalities. The central idea is to convert data from a source modality—such as video \cite{kwon2020imutube, kwon2021approaching, rey2019, lu2022}, text \cite{leng2023generating, leng2024imugpt}, or motion capture \cite{Uhlenberg2022mesh, xiao2021}—into virtual IMU signals. 

 A prominent example of a video-based method is IMUTube \cite{kwon2020imutube, kwon2021approaching}, which converts 2D videos into virtual IMU data. It first extracts 2D and 3D human poses from the videos, then estimates joint rotations and full-body motion, from which virtual IMU signals are generated. This pipeline supports scalable data generation from publicly available video sources such as YouTube or large video datasets like Kinetics-400 \cite{kay2017kineticshumanactionvideo}. The generated virtual IMU data has shown to significantly enhance the performance of downstream HAR models, particularly for activities involving large-scale movements such as locomotion or gym exercises. However, follow-up evaluations revealed that the benefits are more limited for activities involving subtle motions, such as driving \cite{Leng2023imutube}.

In parallel, text-based approaches such as IMUGPT \cite{leng2023generating, leng2024imugpt} leverage LLMs to generate diverse textual descriptions of daily activities. Using a text-to-motion model, these textual descriptions are then converted into 3D motion sequences and subsequently into virtual IMU signals. The diversity in text prompts aims to reflect the wide range of ways in which humans perform activities in real life. Training models on this diverse set of virtual IMU signals can lead to improved generalizability and robustness in activity recognition.

Both video- and text-based approaches have shown promise in improving model performance, especially in scenarios where real IMU data is scarce or imbalanced. However, existing evaluations have examined these modalities in isolation, making direct comparisons difficult. This work provides the first head-to-head evaluation of video- and text-based virtual IMU generation methods under consistent experimental settings, enabling a more rigorous assessment of their relative effectiveness.

\subsection{Data Augmentation}

Data augmentation is a widely used strategy to improve model generalization by synthetically increasing the diversity of the training data. Whereas in computer vision common transformations include cropping, rotation, flipping, and color jittering \cite{shorten2019survey}, sensor-based HAR has adopted domain-specific augmentation techniques designed to perturb and thus enhance sensor signals.

Typical augmentations for IMU signals include axis permutation, adding Gaussian noise, temporal warping, signal scaling, and rotation transformations. These methods introduce controlled variations to the input signals and have been shown to improve model robustness to inter-subject and intra-subject variability \cite{munzner2017cnn, fawaz2018dataaugmentationusingsynthetic, alawneh2021enhancing, choi2023effects}. Beyond these traditional methods, some studies have explored the use of Generative Adversarial Networks (GANs) to synthesize entirely new sensor data as a form of data augmentation \cite{li2020activitygan,hu2023bsdgan} -- yet, overall, only with moderate or no success. 

 In this study, we provide a direct comparison between classical data augmentation and two cross-modality transfer pipelines—IMU\-Tube and IMUGPT—to assess their respective benefits for downstream HAR model performance.

\section{Comparative Study of Cross-Modality Transfer and Data Augmentation}

\begin{figure*}[t]
    \centering
    \includegraphics[width=0.95\linewidth]{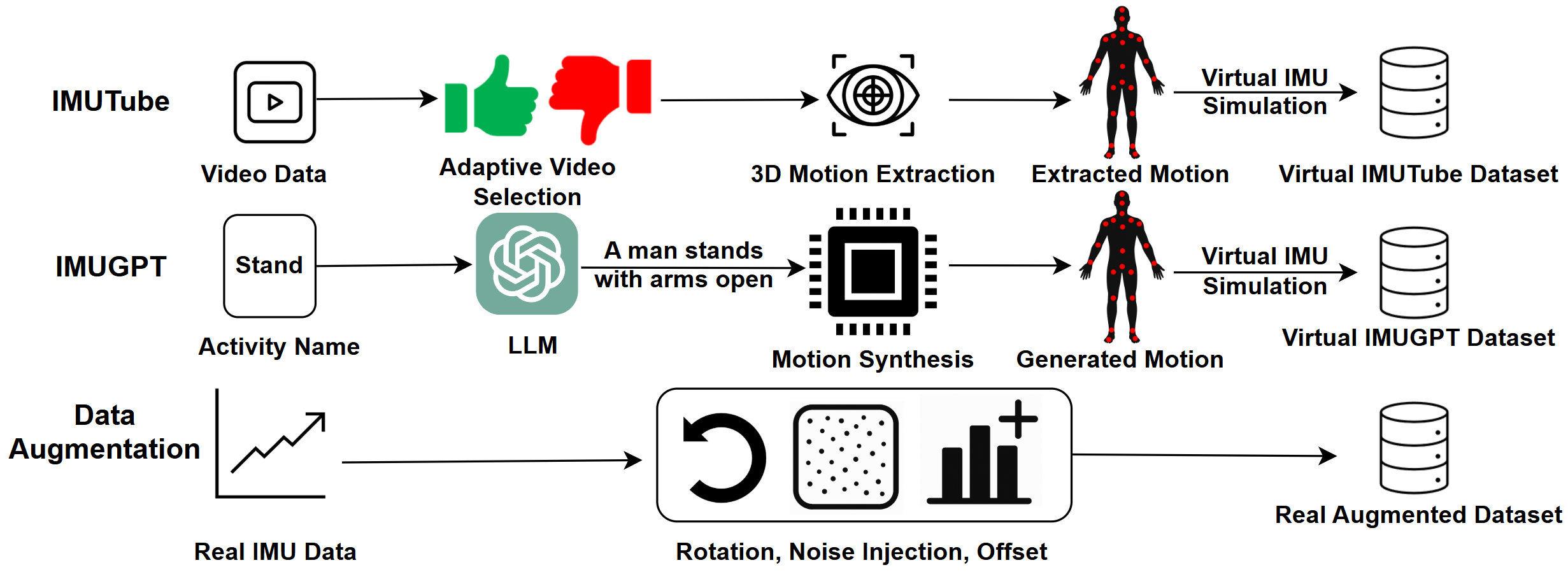}
    % \vspace{-0.1in}
    \caption{An overview of the three ways to scaling data. 
    }
    % \vspace{-0.2in}
    \label{fig:flowchart}
\end{figure*}

This paper presents a comparative study designed to evaluate three approaches for expanding training data in HAR: (1) text-based cross-modality transfer (IMUGPT); (2) video-based cross-modality transfer (IMUTube); and (3) classical sensor-level data augmentation. We  generate large-scale virtual IMU datasets using (1) and (2) and compare their effectiveness against traditional augmentation techniques by training and evaluating four HAR models across three real-world datasets. \autoref{fig:flowchart} gives an overview of our study.

% \vspace{-0.15in}

\subsection{Large Scale Virtual IMU Data Generation}

To construct our large-scale labeled virtual IMU dataset, we leveraged the Kinetics-400 dataset \cite{kay2017kineticshumanactionvideo} as our source of reference videos. Kinetics-400 contains 400 diverse human action classes, each represented by 200 to 700 video clips, which was originally collected to support training of video-based human action recognition models.

From the Kinetics-400 dataset, we selected a subset of 100 action classes to serve as the basis for our virtual IMU dataset, aiming to capture a diverse range of human activities while staying within our computational constraints.
To ensure a representative and balanced selection, we first grouped the 400 classes into broader semantic categories such as sports, daily activities, occupational tasks, and social interactions. Within each category, we then selected action classes that maximized overlap with activities commonly found in existing HAR datasets. This approach enabled us to focus on classes that are both semantically meaningful and relevant for sensor-based recognition. 

To generate virtual IMU data for these activities, we employed two  approaches: one based on textual descriptions and the other based on video inputs (details given below). We then compare the utility of the generated data from both approaches to improve the performance of downstream HAR models. 

\subsubsection{Text-Based Virtual IMU Data Generation}

To generate text-based virtual IMU data, we adopted the IMUGPT pipeline \cite{leng2024imugpt,leng2023generating}, which can autonomously generate virtual IMU data by automatically generating natural language descriptions of human activities, and then converts the textual descriptions into 3D motion and, subsequently, virtual IMU data. The core idea is to capture the inherent variability in how activities are performed by first generating a diverse set of textual descriptions for each activity class, and then converting those into motion sequences.

For each activity, we used GPT-4o \cite{openai2023gpt4} to generate 500 short, varied descriptions ($\leq$ 15 words) that depict how the activity might be performed. The textual description generation process was guided by a few-shot prompting strategy, where we provided the LLM with example descriptions and high-level instructions (e.g., \texttt{``the activity should involve only a single person, with minimal environmental detail''}). 

The generated textual descriptions were then passed to the motion synthesis model T2M-GPT \cite{zhang2023generating} to produce 3D human motion sequences. T2M-GPT operates by first embedding the text using CLIP \cite{Radford2021clip}, then using the embedding, a transformer autoregressively generating code indices, which are subsequently decoded into latent vectors via a learned codebook. A learned decoder maps these latent representations into motion sequences consisting of 3D positions for 22 human body joints. Each generated textual description gets converted to a motion sequence (between five and ten seconds long), where the exact length depends on when the end token is generated by the transformer. 

We estimated joint rotations using inverse kinematics \cite{Yamane2003ik}, with the pelvis as the root joint following conventions in 3D human pose estimation. Using these rotations and the root joint's translation, we simulated virtual IMU signals using IMUSim \cite{Yound2011imusim}. IMUSim computes linear acceleration and angular velocity for each joint, and introduces realistic sensor noise. Doing so, we generated virtual IMU data for 22 selected sensor locations across the body as shown in \autoref{fig:flowchart}. We refer to the resulting virtual IMU dataset as the Virtual IMUGPT dataset.

\subsubsection{Video-Based Virtual IMU Data Generation}
To generate video-based virtual IMU data, we utilized the IMUTube pipeline \cite{kwon2020imutube, kwon2019handling}. For each selected activity class, we sourced corresponding video clips from the Kinetics-400 dataset and processed them through IMUTube. Because many clips contain segments with occlusions, tracking errors, or unstable camera motion, we first applied an adaptive video selection module \cite{kwon2021approaching} to filter out low-quality segments. This module automatically removes sequences with irrelevant content, distorted or incomplete poses, obstructions, or significant foreground/background movement, ensuring that only segments suitable for pose tracking are retained.

The filtered video segments are then passed to the main IMUTube pipeline. First, 2D human poses are extracted from each frame using ViTPose \cite{xu2022vitpose}, a transformer-based 2D pose estimation model. These 2D poses are then lifted into 3D using MixSTE \cite{Zhang2022mixste}, a state-of-the-art 3D pose estimation model. Next, camera ego-motion is estimated via 3D scene reconstruction techniques \cite{zhao2020towards}. Combining the reconstructed 3D poses and estimated camera motion, IMUTube recovers the subject’s global movement trajectory in space.

As with IMUGPT, we use IMUSim \cite{Yound2011imusim} to simulate virtual IMU sensor readings. Virtual IMU data is extracted from 22 specified joint locations of the human body, as illustrated in \autoref{fig:flowchart}. From here on, we will refer to this generated dataset as the \textit{Virtual IMUTube dataset}.

\subsection{Sensor-Level Data Augmentation}
For a given segment of sensor data, we apply three forms of sensor-level data augmentation. Each segment is represented as a matrix \( \mathbf{X} \in \mathbb{R}^{T \times C} \), where \( T \) is the number of samples in the segment and \( C \) is the total number of channels. Each tri-axial sensor contributes 3 channels, i.e., \( C = 3S \), where \( S \) is the number of sensors. The three forms of data augmentation that we applied are shown as follows:

\begin{enumerate}
    \item \textbf{Rotation Augmentation.}
    For each sensor \( s \in \{1, \dots, S\} \), we extract the corresponding tri-axial data \( \mathbf{X}^{(s)} \in \mathbb{R}^{T \times 3} \) and apply a fixed rotation around the z-axis. The rotation matrix is defined as:
    \[
    \mathbf{R}_z = 
    \begin{bmatrix}
        \cos(\theta) & -\sin(\theta) & 0 \\
        \sin(\theta) & \cos(\theta) & 0 \\
        0 & 0 & 1
    \end{bmatrix}, \quad \text{with } \theta = \frac{\pi}{6}.
    \]
    The rotated signal for each sensor is then computed as:
    \[
    \mathbf{X}^{(s)}_{\text{rot}} = \mathbf{X}^{(s)} \cdot \mathbf{R}_z^\top.
    \]
    All rotated sensor signals are concatenated to form \( \mathbf{X}_{\text{rot}} \in \mathbb{R}^{T \times C} \).

    \item \textbf{Gaussian Noise Injection.}
    We add element-wise Gaussian noise sampled from \( \mathcal{N}(0, \sigma^2) \) with \( \sigma = 0.05 \), resulting in:
    \[
    \mathbf{X}_{\text{noise}} = \mathbf{X} + \boldsymbol{\epsilon}, \quad \boldsymbol{\epsilon} \sim \mathcal{N}(0, 0.05^2).
    \]

    \item \textbf{Sensor Bias Simulation.}
    To emulate calibration drift or bias, we add a constant offset \( \mathbf{b} \in \mathbb{R}^{1 \times C} \), where each component is sampled from the uniform distribution \( \mathcal{U}(-0.1, 0.1) \). The bias-augmented signal is:
    \[
    \mathbf{X}_{\text{bias}} = \mathbf{X} + \mathbf{1}_{T} \cdot \mathbf{b},
    \]
    where \( \mathbf{1}_{T} \in \mathbb{R}^{T \times 1} \) is a vector of ones to broadcast the bias across all timesteps.
\end{enumerate}

For a given real IMU dataset, augmentation is applied to all segmented sensor windows obtained through a sliding window approach. The final training dataset includes \( \mathbf{X} \), \( \mathbf{X}_{\text{rot}} \), \( \mathbf{X}_{\text{noise}} \), and \( \mathbf{X}_{\text{bias}} \), which are concatenated along the batch dimension for training.

\subsection{Datasets}
% \vspace{-0.05in}
\begin{table}[t]
\centering
\caption{Training dataset sizes (in minutes). \textit{IMUGPT} and \textit{IMUTube} refer to subsets of the virtual datasets containing only activities overlapping with each real dataset. \textit{Real Augmented} denotes the size after applying  data augmentation. }
\vspace{-0.1in}
\begin{adjustbox}{width=0.45\textwidth}
\begin{tabular}{lllll}
\hline
Datasets & Real & IMUGPT & IMUTube & Real Augmented \\
\hline
UTD-MHAD & 21 & 544 & 196 & 85 \\
PAMAP2   & 53 & 272 & 42  & 212 \\
HAD-AW   & 399 & 943 & 255 & 1,595 \\
\hline
\end{tabular}
\end{adjustbox}
\label{table:dataset}
% \vspace{-0.05in}
\end{table}

We conducted our experiments using three publicly available HAR datasets: UTD-MHAD \cite{chen2015mhad}, PAMAP2 \cite{Reisspamap}, and HAD-AW \cite{Mohammed2018hadaw}. The UTD-MHAD dataset comprises recordings from eight subjects performing 27 activities, with each activity repeated four times per subject. IMU sensors were positioned on the right thigh for six lower-body activities and on the right wrist for the remaining 21 upper-body activities. The PAMAP2 dataset includes data from 9 subjects executing 18 different activities while wearing three IMU sensors placed on the chest, dominant wrist, and dominant ankle. The HAD-AW dataset contains recordings from 16 subjects who each performed 31 activities ten times while wearing an Apple Watch Series One on their right wrist.

For each dataset, we selected only the activities that overlapped with those available in the Kinetics-400 dataset, ensuring that the generated virtual IMU datasets contains data for these activities. The final set of  activities is listed in \autoref{table:activities}. We downsampled all real IMU recordings to 20 Hz to match the sampling rate of the virtual IMU data. 

% \vspace{-0.05in}
\begin{table}[t]
\centering
\caption{Selected activity classes from each dataset that overlap with those available in Kinetics-400.}
\renewcommand{\arraystretch}{1.2}
\begin{tabular}{c|p{6.2cm}}
\toprule
\textbf{Dataset} & \textbf{Selected Activity Classes} \\
\midrule
\textbf{HAD-AW} & Riding a bike, Dancing ballet, Drawing, Driving car, Eating burger, Playing piano, Playing guitar, Jogging, Using computer, Washing hands, Writing, Playing violin, Taking a shower, Making bed \\
\hline
\textbf{PAMAP2} & Walking, Jogging, Biking, Using computer, Folding clothes \\
\hline
\textbf{UTD-MHAD} & Clapping, Throwing ball, Shooting basketball, Drawing, Boxing, Hitting baseball, Hitting tennis ball, Push, Jogging, Walking, Squat \\
\bottomrule
\end{tabular}
\label{table:activities}
\end{table}

\subsection{Experimental Settings}
\subsubsection{Classification Models}
We use four commonly adopted models in the HAR research community for our study: a Random Forest classifier, DeepConvLSTM \cite{convlstm}, DeepConvLSTM with self-attention \cite{singh2021DeepConvLSTM} and Bi-LSTM with attention \cite{zhang2023bilstm}. To segment the IMU data, we apply a sliding window approach with a window size of two seconds and a one-second overlap between adjacent segments. The Random Forest model is trained on ECDF features  \cite{hammerla2013preserving}, comprising 15 components extracted from the windows. DeepConvLSTM, DeepConvLSTM with self-attention,  and BiLSTM with attention are trained directly on raw IMU signals for up to 30 epochs using the Adam optimizer, with learning rates dynamically adjusted via the ReduceLROnPlateau scheduler \cite{reducelronplateau}. 

Hyperparameters, including learning rate (ranging from $10^{-6}$ to $10^{-2}$) and weight decay (from $10^{-4}$ to $10^{-3}$), are selected through grid search. For the UTD-MHAD and PAMAP datasets, we conduct leave-one-subject-out cross-validation for all three models. For the HAD-AW dataset, we instead employ 5-fold stratified cross-validation, as not all subjects performed the full set of activities in the released dataset. Cross-validation for each model is repeated three times using different random seeds, and we report the average macro F1 score along with the standard deviation across these runs.

\subsubsection{Training Configurations}
We evaluate the following training configurations:

\begin{enumerate}
\item \textbf{Real Only:} The model is trained exclusively on the real IMU training dataset.

\item \textbf{Real + IMUGPT:} The training set consists of the real IMU training dataset combined with a subset of the Virtual IMU\-GPT dataset. The subset includes only data samples corresponding to the same activities and sensor locations as in the real dataset.

\item \textbf{Real + IMUTube:} Similar to the above, this configuration combines the real IMU training dataset with a subset of the Virtual IMUTube dataset, matched by activity types and sensor placements.

\item \textbf{Real + IMUGPT + IMUTube:} The model is trained on a combined dataset consisting of the real IMU data, along with subsets of both the Virtual IMUGPT and Virtual IMUTube datasets aligned in terms of activities and sensor locations.

\item \textbf{Real + Augmentation:} The training dataset includes only the real IMU training data, with data augmentation techniques applied.
\end{enumerate}

For each training configuration, we additionally conducted experiments using only 10\% of the real IMU training data to simulate scenarios where access to labeled IMU data is limited. In all cases, evaluation was performed on the unmodified real IMU test sets.

\begin{table}[t]
  \centering
  \caption{Comparison of Model Performance (Macro F1) using Full Real Data. Green shading indicates the best training configuration for each dataset and model combination.}
  \small
  \begin{tabular}{l|c|c|c}
     & UTD-MHAD & PAMAP2 & HAD-AW \\
    \toprule
    \multicolumn{4}{c}{Random Forest Classifier} \\
    \midrule
    Real Only               & 65.26 ± 0.34  & 48.09 ± 1.68   & \cellcolor[HTML]{caebc0}55.68 ± 0.22 \\
    Real + IMUGPT           & 63.41 ± 0.49  & 66.22 ± 2.34   & 53.19 ± 0.07        \\
    Real + IMUTube          & 66.65 ± 0.41  & 86.46 ± 2.38   & 54.64 ± 0.10        \\
    Real + IMUGPT + IMUTube & 65.06 ± 0.51  & \cellcolor[HTML]{caebc0}88.35 ± 2.28   & 52.94 ± 0.27        \\
    Real + Augmentation     & \cellcolor[HTML]{caebc0}67.37 ± 0.62  & 47.60 ± 1.78   & 55.34 ± 0.19        \\
    \midrule

    \multicolumn{4}{c}{DeepConvLSTM} \\
    \midrule
    Real Only               & 51.36 ± 1.19  & 48.47 ± 9.11   & 66.53 ± 0.20        \\
    Real + IMUGPT           & 52.47 ± 0.29  & 74.01 ± 0.54   & 64.54 ± 0.16        \\
    Real + IMUTube          & 54.23 ± 0.21  & 82.48 ± 3.18   & 65.18 ± 0.37        \\
    Real + IMUGPT + IMUTube & 52.09 ± 0.26  & \cellcolor[HTML]{caebc0}86.59 ± 0.88   & 63.85 ± 0.36        \\
    Real + Augmentation     & \cellcolor[HTML]{caebc0}57.33 ± 1.00  & 55.84 ± 0.31   & \cellcolor[HTML]{caebc0}67.15 ± 0.33 \\
    \midrule

    \multicolumn{4}{c}{DeepConvLSTM with Self-Attention} \\
    \midrule
    Real Only               & 51.86 ± 0.67  & 71.42 ± 1.81   & \cellcolor[HTML]{caebc0}67.79 ± 0.11 \\
    Real + IMUGPT           & 49.82 ± 0.52  & 84.31 ± 4.74   & 65.63 ± 0.23        \\
    Real + IMUTube          & 57.37 ± 0.46  & 82.64 ± 2.11   & 65.94 ± 0.30        \\
    Real + IMUGPT + IMUTube & 49.30 ± 0.54  & \cellcolor[HTML]{caebc0}90.29 ± 0.65   & 64.65 ± 0.23        \\
    Real + Augmentation     & \cellcolor[HTML]{caebc0}58.26 ± 1.69  & 62.34 ± 4.71   & 67.71 ± 0.36        \\
    \midrule

    \multicolumn{4}{c}{BiLSTM with Attention} \\
    \midrule
    Real Only               & 78.57 ± 0.18  & 31.07 ± 2.43   & 89.49 ± 0.25        \\
    Real + IMUGPT           & 60.09 ± 1.49  & 47.90 ± 8.70   & 82.70 ± 0.28        \\
    Real + IMUTube          & \cellcolor[HTML]{caebc0}84.51 ± 1.99  & 52.39 ± 6.53   & 81.19 ± 0.30        \\
    Real + IMUGPT + IMUTube & 64.33 ± 7.17  & \cellcolor[HTML]{caebc0}69.37 ± 9.88   & 81.28 ± 0.27        \\
    Real + Augmentation     & 78.82 ± 0.78  & 35.58 ± 3.81   & \cellcolor[HTML]{caebc0}89.51 ± 0.22 \\
    \midrule
  \end{tabular}
  \label{tab:f1_full}
\end{table}

\begin{table}[t]
  \centering
  \caption{Comparison of Model Performance (Macro F1) using 10\% of the Real Data. Green shading indicates the best training configuration per dataset.}
  \small
  \begin{tabular}{l|c|c|c}
     & UTD-MHAD & PAMAP2 & HAD-AW \\
    \toprule

    \multicolumn{4}{c}{Random Forest Classifier} \\
    \midrule
    Real Only               & 36.13 ± 1.16   & 42.09 ± 4.89   & 33.88 ± 0.49   \\
    Real + IMUGPT           & 50.64 ± 0.38   & 77.22 ± 1.44   & 36.29 ± 0.37   \\
    Real + IMUTube          & \cellcolor[HTML]{caebc0}61.48 ± 0.69   & 85.30 ± 0.15   & 37.45 ± 0.37   \\
    Real + IMUGPT + IMUTUBE & 60.65 ± 0.58   & \cellcolor[HTML]{caebc0}87.92 ± 2.01   & \cellcolor[HTML]{caebc0}39.23 ± 0.18   \\
    Real + Augmentation     & 53.82 ± 0.65   & 51.86 ± 7.35   & 33.80 ± 0.21   \\
    \midrule

    \multicolumn{4}{c}{DeepConvLSTM} \\
    \midrule
    Real Only               & 12.95 ± 1.19  & 28.72 ± 6.05   & 31.35 ± 0.05   \\
    Real + IMUGPT           & 30.71 ± 0.78  & 77.24 ± 2.27   & 34.58 ± 0.47   \\
    Real + IMUTube          & 36.40 ± 0.24  & 78.93 ± 2.02   & 37.86 ± 0.16   \\
    Real + IMUGPT + IMUTube & \cellcolor[HTML]{caebc0}37.21 ± 0.44  & \cellcolor[HTML]{caebc0}91.08 ± 1.55   & \cellcolor[HTML]{caebc0}38.50 ± 0.38   \\
    Real + Augmentation     & 15.57 ± 4.12  & 39.57 ± 2.57   & 31.05 ± 0.04   \\
    \midrule

    \multicolumn{4}{c}{DeepConvLSTM with Self-Attention} \\
    \midrule
    Real Only               & 19.24 ± 0.22  & 37.34 ± 9.82   & 37.02 ± 0.35   \\
    Real + IMUGPT           & 29.39 ± 1.15  & 86.36 ± 3.16   & 36.88 ± 0.33   \\
    Real + IMUTube          & \cellcolor[HTML]{caebc0}40.92 ± 0.57  & 87.88 ± 0.16   & \cellcolor[HTML]{caebc0}40.52 ± 0.09   \\
    Real + IMUGPT + IMUTube & 33.92 ± 0.41  & \cellcolor[HTML]{caebc0}90.27 ± 0.58   & 38.99 ± 0.19   \\
    Real + Augmentation     & 23.05 ± 1.66  & 54.07 ± 6.32   & 36.80 ± 0.48   \\
    \midrule

    \multicolumn{4}{c}{BiLSTM with Attention} \\
    \midrule
    Real Only               & 9.47 ± 0.90   & 20.90 ± 6.29  & 51.53 ± 0.30   \\
    Real + IMUGPT           & 12.61 ± 2.23  & 50.58 ± 3.02  & 45.81 ± 0.22   \\
    Real + IMUTube          & \cellcolor[HTML]{caebc0}37.42 ± 0.42  & \cellcolor[HTML]{caebc0}66.79 ± 7.16  & \cellcolor[HTML]{caebc0}52.88 ± 0.20   \\
    Real + IMUGPT+IMUTUBE   & 14.42 ± 1.22  & 59.64 ± 2.35  & 45.41 ± 0.25   \\
    Real + Augmentation     & 11.23 ± 0.55  & 28.43 ± 6.54  & 47.93 ± 0.28   \\
    \midrule
  \end{tabular}
  \label{tab:f1_10}
\end{table}

\subsection{Results}

\autoref{tab:f1_full} shows the model performances for the various training configurations when the full real IMU training dataset is used. When using the full real dataset, we find that incorporating the virtual IMU data-—especially using the virtual IMUTube dataset or the combination of both virtual IMUTube and virtual IMUGPT datasets—-generally leads to the most improvement in downstream model performance, particularly on PAMAP2 and UTD-MHAD.

\begin{enumerate}
    \item \textbf{PAMAP2:} The combination of both virtual IMUTube and IMUGPT datasets yielded the most consistent performance gains. The DeepConvLSTM with Self-Attention model trained on Real + IMUGPT + IMUTube achieved the best macro F1 score of 90.29\%, representing relative improvements of 26.4\%, 44.83\%, 7.1\%, and 9.3\% over training on Real Only, Real + Augmentation, Real + IMUGPT, and Real + IMUTube, respectively.
    
    \item \textbf{UTD-MHAD:} For the random forest classifier and DeepConvLSTM variants, training on Real + IMUTube and Real + Augmentation produced comparable performance. However, for the BiLSTM with Attention model, training on Real + IMUTube achieved the highest macro F1 score of 84.51\%, with relative improvements of 7.6\%, 7.2\%, 40.6\%, and 31.4\% over Real Only, Real + Augmentation, Real + IMUGPT, and Real + IMUGPT + IMUTube, respectively.
    
    \item \textbf{HAD-AW:} Performance was relatively stable across all training configurations. The addition of virtual IMU data or the application of data augmentation techniques did not yield significant improvements over using the unmodified real IMU data alone.
\end{enumerate}

\begin{figure*}[t]
    \centering
    \includegraphics[width=0.69\linewidth]{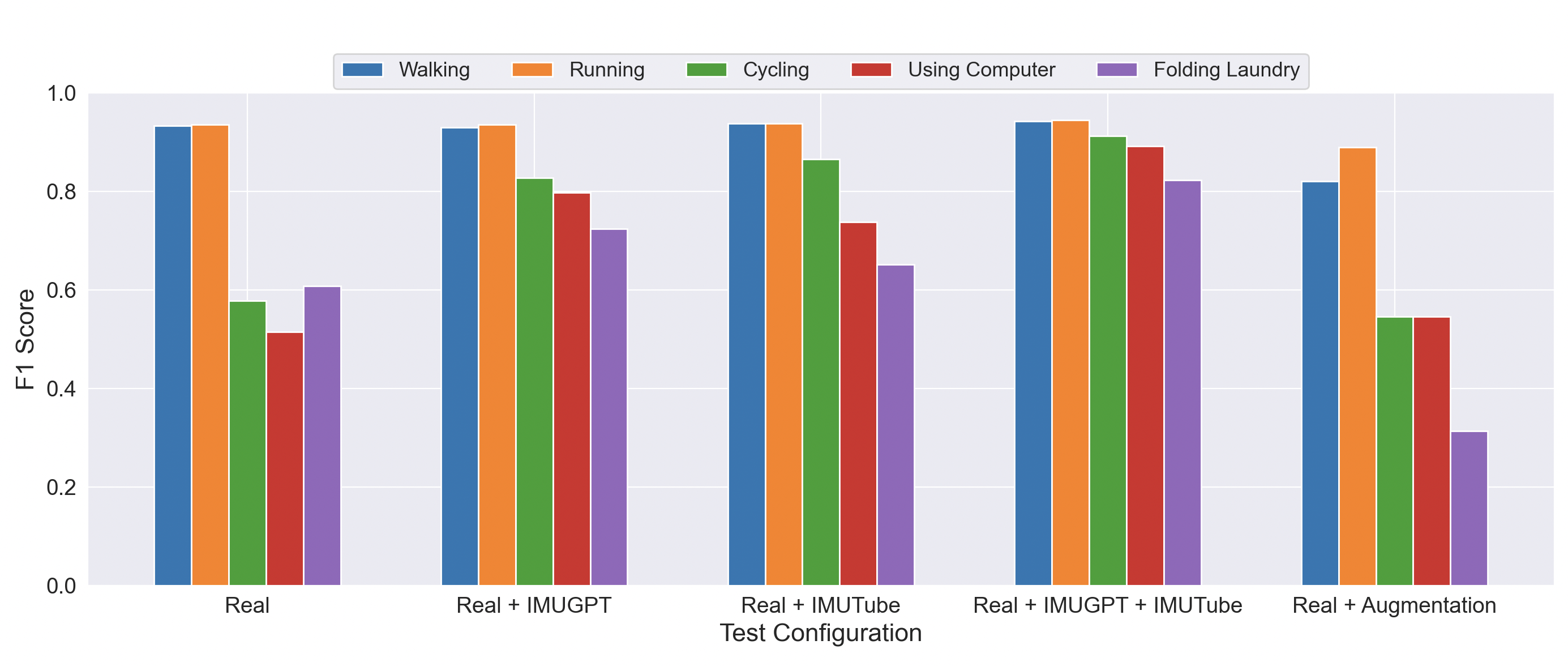}
    \vspace{-0.15in}
    \caption{Per-class macro F1 scores on the PAMAP2 dataset for the DeepConvLSTM with self-attention model under different training configurations.}
    % \vspace{-0.15in}
    
    \label{fig:per_cls_f1}
\end{figure*}

\autoref{tab:f1_10} summarizes the model performance across different training configurations when only 10\% of the real IMU training dataset is used. Overall, incorporating virtual IMU data—particularly from the IMUTube dataset or using both virtual IMU datasets—leads to the most improvements across the models and datasets.

\begin{enumerate}
    \item \textbf{PAMAP2:} The best performance was consistently achieved by models trained on Real + IMUGPT + IMUTube. For instance, the DeepConvLSTM model trained on this configuration reached the highest macro F1 score of 91.08\%, showing relative improvements of 217.1\%, 132.1\%, 17.9\%, and 15.4\% over Real Only, Real + Augmentation, Real + IMUGPT, and Real + IMUTube, respectively.
    
    \item \textbf{UTD-MHAD:} Although the overall scores were lower due to the reduced real IMU training data, models trained with the virtual IMUTube dataset—either alone or in combination with virtual IMUGPT data—still outperformed other configurations. The Random Forest classifier trained on Real + IMUTube achieved the highest macro F1 score of 61.48\%, representing relative improvements of 70.2\%, 14.2\%, 21.4\%, and 1.4\% over Real Only, Real + Augmentation, Real + IMUGPT, and Real + IMUGPT + IMUTube, respectively.
    
    \item \textbf{HAD-AW:} Performance gains were more modest but consistent. Configurations using Real + IMUGPT + IMUTube or Real + IMUTube achieved the highest macro F1 scores across most models, suggesting that virtual IMU data becomes especially beneficial when real data is scarce.
\end{enumerate}

\section{Discussion}
Across both full and limited (10\%) real data training regimes, all three data expansion strategies-—IMUTube, IMUGPT, and classical data augmentation-—were evaluated against models trained on real IMU data only. Virtual IMU data generated via IMUTube and IMUGPT consistently led to substantial improvements in model performance across datasets and training conditions. In contrast, classical data augmentation produced mixed results and did not consistently outperform real-only baselines, highlighting the limitations of perturbation-based techniques in expanding data diversity.

\paragraph{Cross-Modality vs.\ Classical Augmentation.} Cross-modality transfer techniques generally yielded larger performance gains than classical sensor-level data augmentation. On PAMAP2, for instance, combining virtual IMU data from both IMUGPT and IMUTube led to a 78\% relative improvement in macro F1 over the real-only baseline using full data, compared to just a 4\% gain with classical augmentation. In low-data conditions (10\% real data), this performance gap widened further, with IMUGPT+IMUTube achieving 163.3\% improvement versus 35.5\% with augmentation. These results suggest that \textbf{cross-modality methods provide richer data diversity and semantics beyond what can be achieved by simple perturbations of existing signals.}

\paragraph{IMUTube vs.\ IMUGPT.} Between the two virtual IMU data sources, IMUTube consistently delivered slightly better performance than IMUGPT, particularly on datasets with more complex physical activities. For example, on PAMAP2 with full data, Real+IMUTube provided a 58.08\% gain versus 40.65\% for Real+IMU\-GPT. This could be because video-based pose estimation can capture detailed and realistic motion from videos compared to text-based motion synthesis. Still, \textbf{using both IMUGPT and IMUTube together generally gave better results than using either one alone, showing that they complement each other-—IMUGPT enriches diversity through diverse generated textual descriptions, while IMUTube provides accurate movement patterns.}

\paragraph{Data Availability.} While all data expansion methods were beneficial under both full and limited real data regimes, their relative effectiveness was magnified in the 10\% condition. Top-performing configurations achieved an average 116.9\% boost over the Real Only baseline with 10\% of the data, compared to a 28.8\% improvement under full data. This underscores \textbf{the value of additional training data when the amount of real IMU data is limited.} 

\paragraph{Class Specific Improvement.}In addition to overall improvements, synthetic data notably enhanced performance on specific activity classes. Figure~\ref{fig:per_cls_f1} presents per-class F1 scores for PAMAP2 across training configurations. We observed substantial gains for activities such as \textit{biking}, \textit{using computer}, and \textit{folding clothes}, where real-only models performed poorly. These tasks involve motions that may be harder to generalize from limited real data. The addition of virtual IMU signals—particularly from IMUGPT, which introduces diverse activity descriptions—helped improve performance on these classes, suggesting that \textbf{virtual IMU data can help mitigate class imbalance or under-representation.}

\paragraph{Computational Cost} From a deployment perspective, the computational cost of each method varies significantly. Classical data augmentation is computationally negligible. On Nvidia A6000, IMUGPT requires approximately 10 seconds to generate 10 seconds of virtual IMU data. In contrast, IMUTube takes approximately 5 minutes for the same output duration. Given these trade-offs, \textbf{we recommend IMUGPT as a practical starting point for practitioners with limited computational or video resources. For those with access to greater resources, combining IMUTube and IMUGPT provides the most benefit and yields the most robust models.}

\section{Conclusion}

This paper presented a comparative study between classical data augmentation and two cross-modality transfer methods—IMUGPT and IMUTube—for virtual IMU data generation in HAR. We generated a large virtual IMU dataset spanning 100 activities and evaluated these methods across three datasets and four models. Results show that virtual IMU data consistently outperforms real-only and augmented baselines, especially in low-data regimes. IMUTube provides more accurate movement information, while IMUGPT contributes to broader data diversity; combining both often yields the best performance. Our findings offer practical guidance for selecting data generation strategies and highlight the value of virtual IMU data in scaling HAR. 

\section*{Acknowledgments}
This work was partially supported by the NSF Research Fellowship under Grant No. DGE-2039655.

Any opinion, findings, and conclusions or recommendations expressed in this material are thoseof the authors(s) and do not necessarily reflect the views of the National Science Foundation.

\bibliographystyle{ACM-Reference-Format}
\bibliography{./bibs/har}

\end{document}